\documentclass{article} % For LaTeX2e
\usepackage{arxiv,times}
\usepackage{booktabs}
\usepackage{multirow}
\usepackage{amsmath,amsfonts,bm}

\def\eqref#1{equation~\ref{#1}}
\def\1{\bm{1}}

\DeclareMathAlphabet{\mathsfit}{\encodingdefault}{\sfdefault}{m}{sl}
\SetMathAlphabet{\mathsfit}{bold}{\encodingdefault}{\sfdefault}{bx}{n}

\usepackage{hyperref}
\usepackage{url}
\usepackage[pdftex]{graphicx}

\title{The Program Testing Ability of Large Language Models for Code}

\author{
Weimin Xiong$^{1,2}$, Yiwen Guo$^3$\thanks{Corresponding author}\, , Hao Chen$^4$  \\
$^1$Tencent Security Big Data Lab,\, $^2$Peking University,\, $^3$Independent Researcher,\, $^4$UC Davis\\
\small{\texttt{weiminxiong@tencent.com,\ guoyiwen89@gmail.com,\ chen@ucdavis.edu}}
}

\iclrfinalcopy % Uncomment for camera-ready version, but NOT for submission.
\begin{document}

\maketitle

\begin{abstract}
Recent development of large language models (LLMs) for code like CodeX and CodeT5+ demonstrates tremendous promise in achieving code intelligence. 
Their ability of synthesizing code that completes a program for performing a pre-defined task has been intensively tested and verified on benchmark datasets including HumanEval and MBPP. 
Yet, evaluation of these LLMs from more perspectives (than just program synthesis) is also anticipated, considering their broad scope of applications in software engineering.
In this paper, we explore the ability of LLMs for testing programs/code. By performing thorough analyses of recent LLMs for code in program testing, we show a series of intriguing properties of these models and demonstrate how program testing ability of LLMs can be improved.
Following recent work which utilizes generated test cases to enhance program synthesis, we further leverage our findings in improving the quality of the synthesized programs and show +11.77\% and +4.22\% higher code pass rates on HumanEval+ comparing with the GPT-3.5-turbo baseline and the recent state-of-the-art, respectively. Our code is available at \href{https://github.com/WeiminXiong/TestingLLM}{https://github.com/WeiminXiong/TestingLLM}.
% will be publicly available
\end{abstract}

\section{Introduction}

The community has witnessed a surge in the development of large language models (LLMs), which have achieved incredible ability in understanding and generating not only texts but also code. 
LLMs for code (CodeX~\citep{chen2021evaluating}, StarCoder~\citep{li2023starcoder}, CodeT5+~\citep{wang2023codet5+}, etc) have been widely adopted to a variety of applications to achieve code intelligence. 
However, current evaluation of these LLMs mostly focuses on program completion/synthesis, despite the models can also be utilized in other applications.
As the field continues to advance, evaluation of these models from more perspectives is anticipated, which could facilitate deeper understanding of the LLMs. 

As the core of analyzing the behavior of code, the ability of generating proper test cases is of great desire to software engineering. 
Although embryonic development of using deep models in testing has been shown~\citep{tufano2020unit, tufano2022methods2test}, with the remarkable progress in LLMs, it is unclear how far have such abilities of AI been advanced when these powerful models are equipped. 
In this paper, we, for the first time, analyze the ability of recent LLMs in testing programs/code.
Our analyses are performed based on 164 problems from HumanEval+~\citep{chen2021evaluating} and 427 sanitized problems from MBPP~\citep{austin2021program}. We consider 4 test-case generation settings (i.e., self-generated, all-generated, oracle, and placeholder in Figure~\ref{fig:settings}) and test a collection of 11 competitive LLMs for code (including 4 LLMs that have around 1 billion parameters and 7 substantially larger LLMs). 
% The tested LLMs include decoder-only models and encoder-decoder models. 
We conducted a variety of experiments, from which many intriguing takeaway messages are delivered.
% the following takeaway messages can be inferred from our experimental results. For instance, we show that {\color{red}xxxxx}

Several very recent papers~\citep{shi2022natural, li2023towards, chen2023codet} have shown that appropriate usage of even generated test cases can improve the quality of program synthesise, in a spirit that the synthesized programs that could pass a large number of test cases are more likely to be correct. 
Nevertheless, the quality of the generated test cases largely impacts the performance of such methods.
Due to the lack of systematic evaluation of the testing ability of LLMs for code, it is unclear how to craft test cases that could be potentially more helpful to program synthesis and, more broadly, code intelligence.
The studies in this paper aim to shed light on this.
We will demonstrate that, substantially improved program synthesis performance can be obtained by utilizing takeaway messages in our studies. 
Specifically, on GPT-3.5-turbo, we can achieve +11.77\% higher code pass rate on HumanEval+, in comparison with the GPT-3.5-turbo baseline. When compared with a very recent state-of-the-art called CodeT, our solution achieves +4.22\% higher code pass rate.

\section{Evaluation Metrics}
\label{sec:metrics}

To make the evaluation more reliable and comprehensive, it is crucial to first design some suitable metrics, like BLEU~\citep{papineni2002bleu}, ROUGE~\citep{lin2004rouge}, and the pass rate~\citep{chen2021evaluating} for evaluating machine translation, text summarization, and program synthesis, respectively.
In this section, we specify two main evaluation metrics to evaluate the program testing ability of LLMs, from the perspective of correctness and diversity.

\textbf{Pass rate } In software engineering, we expect test cases to represent some desired ``ground-truth'' functionality of the tested program/code. 
In practice, such ``ground-truth'' functionality can be described in the header comments of a function (i.e., docstrings of the function) and tested using the oracle implementation, as in HumanEval~\citep{chen2021evaluating} and MBPP~\cite{austin2021program}. 
The oracle program/code should be able to pass the test, if a generated test case is correct.
Therefore, we leverage the pass rate as a measure to evaluate the correctness of the generated test cases.
% This metric measures the correctness of the test cases generated by the model. 
For a fair comparison, we instruct each model to generate three test cases in the prompt, and, when a model generates more than three test cases, we select the first three for evaluation. 
Assuming that there are in total $M$ programming problems in an experimental dataset and, for each problem, we have $N$ program/code implementations to be generated test cases for.
Each model has only one chance to generate these test cases for each program/code. 
Then, we calculate the pass rate as:
\begin{equation}\label{eq:pass_rate}
P = \frac{1}{MN}\sum_{i=1}^M\sum_{j=1}^N\frac{p_{ij}}{n_{ij}},
\end{equation}
where $n_{ij}$ is the number of test cases in $\mathcal{Q}_{ij}$ which includes no more than three test cases generated for the $j$-th program/code implementation of the $i$-th problem by the evaluated LLM at once, i.e., $\mathcal{Q}_{ij}=\{(x_{ijk},y_{ijk})\}_k$, and $p_{ij}$ is the number of test cases (in $\mathcal{Q}_{ij}$) that do not fail the oracle.

\begin{figure}[t]
    \centering
    \includegraphics[width=\textwidth]{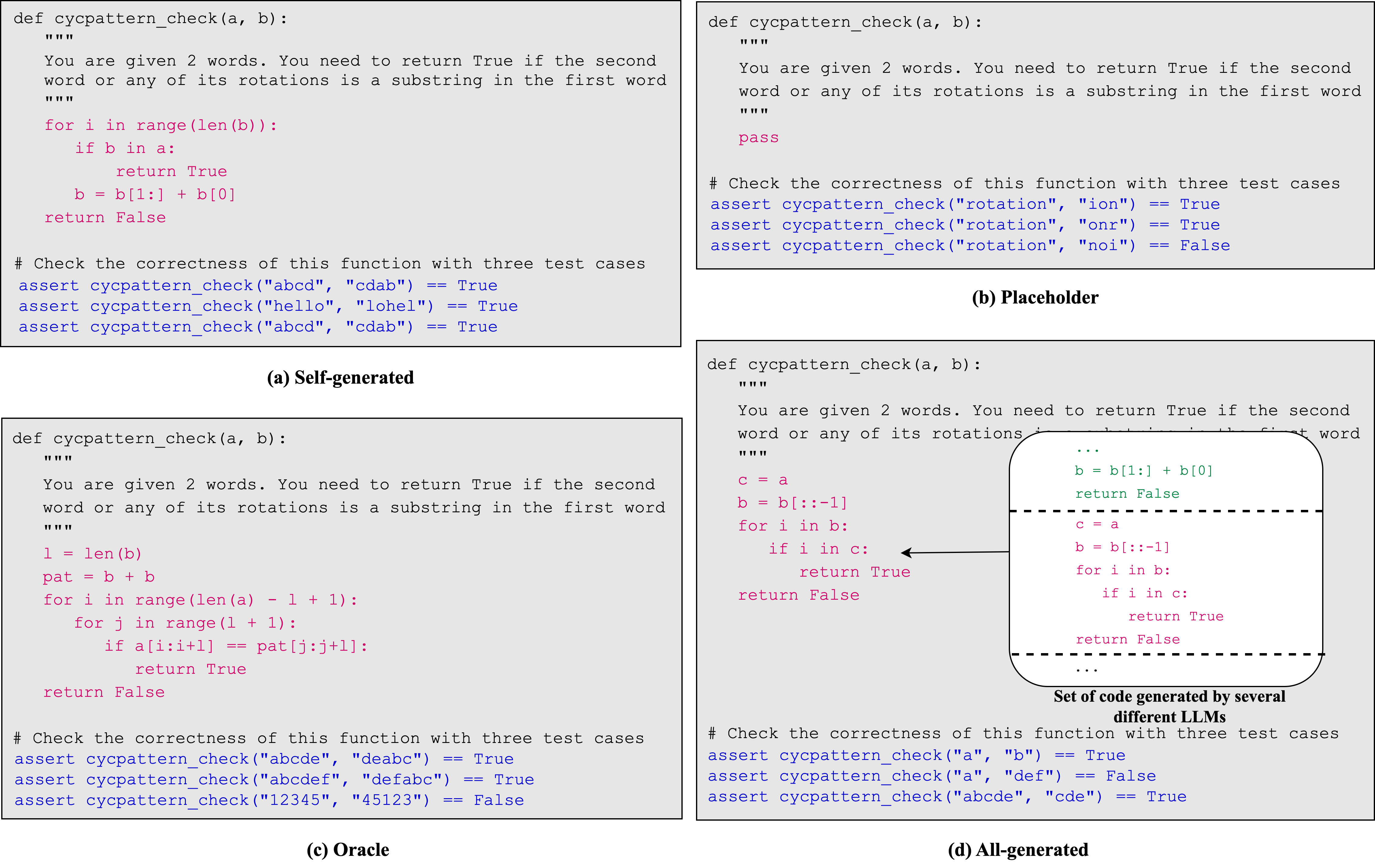} \vskip -0.15in
    \caption{Generating test cases for (a) self-generated code, (b) placeholder, (c), oracle code, and (d) all-generated code.} 
    \label{fig:settings} \vskip -0.2in
\end{figure}

The pass rate defined in Eq.~(\ref{eq:pass_rate}) measures correctness of the generated test cases. 
However, as can be seen in Figure~\ref{fig:settings}, the model can generate duplicate test cases that are less helpful, even though they are correct. 
To avoid such an evaluation bias, we further advocate deduplication in the set of test cases that are considered as correct, which leads to computation of a deduplicated pass rate defined as $P' = \frac{1}{MN}\sum\sum p'_{ij}/n'_{ij}$, where we use $'$ to denote the numbers of unique test cases.

\textbf{Coverage rate } In addition to the above pass rates, we further consider coverage rate as a more fine-grained metric for evaluating the diversity of the generated test cases. 
According to its definition, converge rate computes the degree to which the code is executed, given a test case. 
Since, for each program/code, we keep no more than three test cases at once, we calculate how much percentage of the control structure is covered given these test cases.
Similar to Eq.~(\ref{eq:pass_rate}), we evaluate the performance of testing all programs/code over all $M\times N$ times of generation, i.e., we calculate 
\begin{equation}
C = \frac{1}{MN}\sum_{i=1}^M\sum_{j=1}^N c_{ij}.
\end{equation}
We utilize the \textit{pytest}~\footnote{https://pytest.org} library to evaluate branch coverage for all the three test cases for each code and aggregate the results for all programs/code and all problems.
Apparently, a higher $C$ indicates better testing ability of an LLM, since we expect all parts of the programs/code to be executed to find our all potential bugs, with the set of test cases generated by this LLM.

\section{Large Language Models for Code}
\label{sec:llm_code}

In this section, we outline the evaluated models. We adopt some ``small'' models whose numbers of parameters are around 1B (to be more specific, from 770M to 1.3B in our choices) and some larger models that achieve state-of-the-art performance in the task of program synthesis. 
% We start with the small models first.

For the small models, we use \textbf{InCoder} (1.3B)~\citep{fried2023incoder}, \textbf{CodeGen2} (1B)~\citep{nijkamp2023codegen2}, \textbf{CodeT5+} (770M)~\citep{wang2023codet5+}, and \textbf{SantaCoder} (1.1B)~\citep{allal2023santacoder}.
InCoder is a unified generative model that can perform program/code synthesis as well as code editing, and it combines the strengths of causal language modeling and masked language modeling.
The CodeGen2 model was trained on a deduplicated subset of the Stack v1.1 dataset~\citep{kocetkov2023the}, and its training is formatted with a mixture of objectives for causal language modeling and span corruption.
CodeT5+ is an encoder-decoder model trained on several pre-training tasks including span denoising and two variants of causal language modeling.
SantaCoder was trained on the Python, Java, and JavaScript code in the Stack dataset. %The main model uses Multi Query Attention and Fill-in-the-Middle objective.
% Among the four models, SantaCoder (1.1B) shows the best \textit{code generation performance}, with a Pass@1 of 15.21\% on HumanEval~\citep{chen2021evaluating}, while InCoder (1.3B), CodeGen2 (1B), and CodeT5+ (770M) show 6.99\%, 9.19\%, and 12.95\%, respectively.
The pass rate~\citep{chen2021evaluating} of programs generated by these models is compared in Table~\ref{tab:model pass rate HumanEvalX}. 
When evaluating the (program) pass rate, we let the model generate 200 code implementations for each problem, and we set the temperature to 0.2, 0.6, and 0.8 for calculating pass@1, pass@10, and pass@100, respectively.

As for larger models that achieve state-of-the-art program synthesis performance, we use \textbf{CodeGen2} (16B)~\citep{nijkamp2023codegen2}, 
\textbf{CodeGen-Multi} 
(16B)~\cite{nijkamp2023codegen},
\textbf{CodeGen-Mono} 
(16B)~\cite{nijkamp2023codegen},
\textbf{StarCoder} (15B)~\citep{li2023starcoder}, 
% \textbf{InstructCodeT5+} (16B)~\citep{wang2023codet5+}, 
\textbf{WizardCoder} (15B)~\citep{luo2023wizardcoder}, \textbf{CodeGeeX2} (6B)~\citep{zheng2023codegeex}, and \textbf{GPT-3.5-turbo}.
CodeGen-Multi and CodeGen-Mono are two large models from the first version of CodeGen.
CodeGen-Multi was first trained on the pile dataset~\citep{gao2020pile} and then trained on a subset of the publicly available BigQuery dataset which contains code written in C, C++, Go, Java, JavaScript, and Python.
Based on the 16B CodeGen-Multi model, CodeGen-Mono (16B) was obtained by further tuning on a set of Python code collected from GitHub.
Given a base model that was pre-trained on 1 trillion tokens from the Stack dataset, the 15B StarCoder model was obtained by training it on 35B tokens of Python code.
WizardCoder further empowers StarCoder with instruction tuning, following a similar instruction evolution strategy as in WizardLM~\citep{xu2023wizardlm}. 
% As its name implies, InstructCodeT5+ is also an intruction-tuned LLM for code, and it is based on CodeT5+ (a much larger CodeT5+ than the one introduced in the previous paragraph though), and its instruction tuning data is obtained by letting pre-trained LLMs generate novel tasks and task-related data, including task instructions, task inputs, and their expected outputs. 
CodeGeeX2, the second generation of a multilingual generative model for code, is implemented based on the ChatGLM2 architecture and trained on more code data.
GPT-3.5-turbo is a very capable commercial LLM developed by OpenAI and we accessed it in August, 2023. 
For these large LLMs, we tested pass@1 of all models except GPT-3.5-turbo (whose result can be directly collected from~\citet{liu2023your}'s paper).
By sorting their pass@1 from high to low, they are ranked as: GPT-3.5-turbo (61.7\%), WizardCoder (46.23\%, 15B), CodeGeeX2 (29.97\%, 6B), StarCoder (27.9\%, 15B), CodeGen-Mono (26.15\%, 16B), CodeGen2 (19.33\%, 16B), CodeGen-Multi (15.35\%, 16B).
The ranks on the MBPP dataset are similar. 

\section{Code to be Tested}
\label{sec:code_to_be_tested}

For evaluating the testing ability of LLMs, we need an oracle to express the ground-truth functionality of the tested code. 
Fortunately, current datasets for evaluating program synthesis performance often provide such oracles (see HuamnEval~\citep{chen2021evaluating} and MBPP~\citep{austin2021program}). 
In our experiments, we utilize an amended version of HumanEval called HumanEval+~\citep{liu2023your}, together with MBPP (the sanitized version).
These datasets are established to evaluate basic Python programming performance of LLMs, and they contain 164 and 427 problems, respectively.
% which can be solved by entry-level programmers. 

\subsection{Imperfect Code Implementations}
\label{sec:self_all_generated}
In order to simulate real-world scenarios where the tested code is often buggy, we first adopt synthesized programs/code as the programs/code to be tested, considering that the synthesis of even state-of-the-art LLMs is still imperfect. 
We evaluate the performance of each LLM in testing code that was generated by itself (which is denoted as ``\textbf{Self-generated}'') and code in a set consisting of program completion results of several different LLMs (which is denoted by ``\textbf{All-generated}''). 
That said, the compared LLMs take different code implementations when generating test cases for each programming problem in the self-generated setting. 
Whereas, in the all-generated setting, the same program/code implementations are given to different LLMs for generating test cases for comparison. 
In practice, we apply InCoder (1.3B), CodeGen2 (1B), CodeT5+ (770M), and SantaCoder (1.1B) to construct the all-generated program/code set, while, in the self-generated setting, each LLM first synthesize code and complete a program to fulfill the requirement of each programming problem, and the LLM then generates test cases with the synthesized programs/code in its prompts.
The temperature for all LLMs is uniformly set to 0.2 for synthesizing the programs/code in both settings.
We obtain 100 program/code completions for each problem and we prompt each LLM to generate 3 test cases for every program/code implementation in the self-generated setting, and we sampled 100 implementations from the synthesis results of InCoder (1.3B), CodeGen2 (1B), CodeT5+ (770M), and SantaCoder (1.1B) to form the all-generated code set, i.e., we have $N=100$ for these settings.
% HumanEval contains 164 programming problems for test, and, 
% For each problem, we generated 100 code implementations using each LLM for the ``Self-generated'' setting

We follow the same way of generating code as introduced in the papers of these LLMs. For model without instruction tuning, like InCoder and CodeT5+, we synthesize programs/code using the default prompt given by each programming problem in the test dataset, while, for models that have adopted instruction tuning, e.g., 
% InstructCodeT5+ and 
WizardCoder, we use the recommended prompt in their papers. 

% As we report the pass@1 score for theirs models, we follow the previous works~\citep{chen2021evaluating} to generate samples with the same set of hyper-parameters: temperature=0.2, and top\_p=0.95.
% For generating test cases, we comply with the way of generating codes introduced in the paper corresponding to each LLM. 

\begin{table}[!t]
    \centering
    \small
    \begin{tabular}{c c c c c}
    \toprule
    \textbf{Model} & \textbf{Size} & \textbf{Pass@1} & \textbf{Pass@10} & \textbf{Pass@100}\\
    \midrule
    InCoder & 1.3B & 6.99\%/14.06\% & 14.20\%/34.98\% & 23.76\%/55.34\% \\ 
    CodeGen2 & 1B & 9.19\%/17.50\% & 16.06\%/36.86\% & 25.90\%/59.32\% \\ 
    CodeT5+ & 770M & 12.95\%/28.02\% & 25.09\%/47.69\% & 37.56\%/65.26\% \\
    SantaCoder & 1.1B & 15.21\%/29.42\% & 26.01\%/51.30\% & 43.80\%/69.10\% \\
    \bottomrule
    \end{tabular} \vskip -0.1in
    \caption{\emph{Program synthesis performance} of the \emph{small} LLMs (whose number of parameters is around 1 billion) evaluated on HumanEval+\,/\,MBPP\,(sanitized).} \vskip -0.17in
    \label{tab:model pass rate HumanEvalX}
\end{table}

\subsection{Optimal Code Implementations}
As a reference, we also report the performance of generating accurate and diverse test cases when the written code is perfectly correct, which is achieved by adopting the oracle as the programs/code to be tested (and such a setting is denoted by ``\textbf{Oracle}''). 
Since~\cite{liu2023your} have reported that some oracle code in the HumanEval dataset can be incorrect, we adopt the amended oracle set in HumanEval+ in this setting. We further used the revised oracle code implementations instead of the original ones in evaluating the pass rate (i.e., $P'$) of the generated test cases.
Considering that the public datasets often only provide one oracle implementation for each problem, and to keep the uncertainty of evaluation results consistent, we copy the oracle implementation by 100$\times$ and we prompt to generate 3 test cases for each of these copies. It can be regarded as letting $N=100$, just like in the previous settings in Section~\ref{sec:self_all_generated}.

\subsection{No Implementation}
In certain scenarios, we require test cases before the function/program has been fully implemented, hence we also evaluate in a setting where the main body of a tested function/program is merely a placeholder, as depicted in Figure~\ref{fig:settings}(b). 
This scenario often occurs when the main code has not yet been implemented for a function/program or the test engineer does not want to introduce implementation bias to the LLM when generating test cases for a function/program.
We denote such a setting as ``\textbf{Placeholder}'' in this paper. 
We also let $N=100$, as in the oracle setting.

\section{Test Case Generation}
\label{sec:test_case_gen}

In this section, we introduce how test cases can be generated, when the implementation of a function/program is given as described in Section~\ref{sec:code_to_be_tested}.
In this paper, a desired test case is a pair of input and its expected output for the function/program defined in the context. 
As an example, Figure~\ref{fig:settings} demonstrates some test cases for the programming problem of checking whether the two words satisfy a specific rotation pattern.
To generate test cases, we use the LLMs introduced in Section~\ref{sec:llm_code}.

We wrote extra prompts to instruct the LLMs to generate three test cases for each given code which include docstrings that describe the purpose of this function, as depicted in Figure~\ref{fig:settings}.
Our instruction commands the LLMs (1) to ``check the correctness of this function with three test'' and (2) to start writing test code with an ``\texttt{assert}'' statement and the tested function, which specifies the format of the test cases as input-output pairs that can be parsed. 
For instance, given the example in Figure~\ref{fig:settings}, the extra prompt should be ``\texttt{\# Check the correctness of this function with three test cases \textbackslash n assert cycpattern\_check}''.

We then concatenate the extra prompt with the code and feed the concatenation into each LLM, for extracting test cases from the model output.
The LLM will try to complete the given input by generating one or more ``\texttt{assert}'' statement(s), and we split the generation results into sub-strings, with ``\texttt{assert}'' as the separator. Each sub-string is then considered as a test statement, and we only take the first three statements if there exist more than three statements, as has been introduced in Section~\ref{sec:metrics}. Such a split can be considered as an effective post-processing operation which largely improves the quality of the generated test code, considering that some non-sense code pieces may be generated in the output of the LLMs.
When using HumanEval+ and MBPP, we try removing test cases in the docstrings of the function, if there exist any, just to get rid of the broad hints from the docstrings~\citep{chen2023codet}.
The temperature for generating test cases is kept as 0.2. 
% Not that we remove all example input-output context before generating test cases to avoid exposing real test cases to the language model and to increase the diversity of the generated test cases.

Once obtained, the generated test cases are then compiled, and evaluated for their correctness and diversity to report the pass rate $P'$ and the coverage rate $C$.
When calculating, for each problem and every set of completions generated, we create a temporary folder. 

% \begin{table}[htbp]
%     \centering
%     \small
%     \begin{tabular}{c c c c c}
%     \toprule
%     \textbf{Model} & \textbf{Size} & \textbf{Pass@1} & \textbf{Pass@10} & \textbf{Pass@100}\\
%     \midrule
%     InCoder & 1.3B &  &  & \\ 
%     CodeGen2 & 1B &  &  & \\ 
%     CodeT5+ & 770M &  &  &  \\
%     SantaCoder & 1.1B &  &  &  \\
%     \bottomrule
%     \end{tabular}
%     \caption{Results of \textbf{code generation} task in MBPP.}
%     \label{tab:model pass rate MBPP}
% \end{table}

% When generating test cases, we adopt four different prompt settings as shown in Figure \ref{fig:settings}: self-generated setting, all-generated setting, oracle setting and placeholder setting. In self-generated setting, the model first performs code generation and then generates test samples corresponding to the code. In all-generated setting, we pick the public part from the collection of all model-generated code, and in doing so, we prioritise the correct code. We generate test cases for this part of code to simulate the generation of test cases in real scenarios. In oracle setting, we generate test cases for the standard code provided in the problem. And in placeholder setting, we replace the standard code with placeholder "pass". As with code generation, we adopt necleur sampling with temperature=0.2, and top\_p=0.95.

\section{Main Results for Test Case Generation}
\label{sec:main}

\begin{table}[bp]
    \centering
    \small
    \begin{tabular}{c c c c c c c}
    \toprule
    \textbf{Model} & \textbf{Size} & \textbf{Oracle} &  \textbf{Self-generated} & \textbf{All-generated}  & \textbf{Placeholder} \\ %& \textbf{Pass Rate}\\
    \midrule
    InCoder & 1.3B & 21.31\%\,(61.43\%) & 23.37\%\,(59.36\%) & 22.72\%\,(61.10\%) & 25.19\%\,(62.75\%) \\ %& 9.45\\
    CodeGen2 & 1B & 31.63\%\,(\textbf{71.55\%}) & 30.62\%\,(69.38\%) & 30.93\%\,(69.70\%) & 30.69\%\,(69.00\%) \\ %& 10.82\\
    CodeT5+ & 770M & \textbf{35.43\%}\,(71.45\%) & \textbf{32.34\%}\,(70.45\%) & \textbf{31.49}\%\,(69.75\%) & \textbf{32.67\%}\,(70.67\%) \\ %& 14.83 \\
    SantaCoder & 1.1B & 30.97\%\,({71.46\%}) & 30.43\%\,(\textbf{70.81\%}) & 30.13\%\,(\textbf{70.55\%}) & 30.78\%\,(\textbf{71.24\%}s) \\ %& 17.85\\
    \bottomrule
    \end{tabular}
    \caption{The pass rates (and coverage rate) of the test cases generated on HumanEval+ in different settings for LLMs with around 1 billion parameters.}
    \label{tab:small model accuracy}
\end{table}

\begin{table}[!t]
    \centering
    \small
    \begin{tabular}{c c c c c c}
    \toprule
    \textbf{Model} & \textbf{Size} & \textbf{Oracle} &  \textbf{Self-generated} & \textbf{All-generated}  & \textbf{Placeholder}\\
    \midrule
    % InstructCodeT5+ & 16B & \\
    CodeGen-Multi & 16B & 43.88\%\,(67.91\%) & 41.85\%\,(69.30\%) & 40.38\%\,(66.97\%) & 39.74\%\,(68.28\%) \\
    CodeGen2 & 16B & 46.34\%\,(73.07\%) & 45.44\%\,(73.17\%) & 42.00\%\,(72.45\%) & 42.69\%\,(72.86\%) \\
    CodeGen-Mono & 16B & 49.03\%\,(74.82\%) & 45.73\%\,(73.74\%)  & 43.91\%\,(73.66\%) & 44.92\%\,(73.63\%) \\
    StarCoder & 15B & 55.07\%\,(76.02\%) & 52.52\%\,(72.45\%) & 48.20\%\,(72.30\%)  & 50.58\%\,(74.52\%) \\
    CodeGeeX2 & 6B & 57.03\%\,(74.42\%) & 53.16\%\,(73.55\%)  & 49.28\%\,(70.32\%) & 51.78\%\,(73.08\%) \\
    WizardCoder & 15B & 53.89\%\,(\textbf{77.87\%}) & 55.47\%\,(76.07\%) & 48.02\%\,(\textbf{75.27\%})  & 49.89\%\,(\textbf{75.12\%})\\
    GPT-3.5-turbo & - & \textbf{71.03\%}\,(77.85\%) & \textbf{72.45\%}\,(\textbf{77.24\%}) & \textbf{59.24\%}\,(74.99\%) & \textbf{66.28\%}\,(74.03\%)\\
    \bottomrule
    \end{tabular}
    \caption{The pass rates (and coverage rate) of the test cases generated on HumanEval+ in different settings for LLMs whose parameters are obviously more than 1 billion.} %\vskip -0.1in
    \label{tab:large model accuracy}
\end{table}

The experiment results of small and large LLMs on HumanEval+ can be found in Table \ref{tab:small model accuracy} and Table \ref{tab:large model accuracy}, respectively. Table~\ref{tab:model accuracy mbpp} shows the results on MBPP.
There are several takeaways from these tables.

\begin{figure}[t]
    \centering
    \begin{minipage}{0.35\textwidth}
        \centering
        \includegraphics[width=0.9\linewidth, height=0.21\textheight]{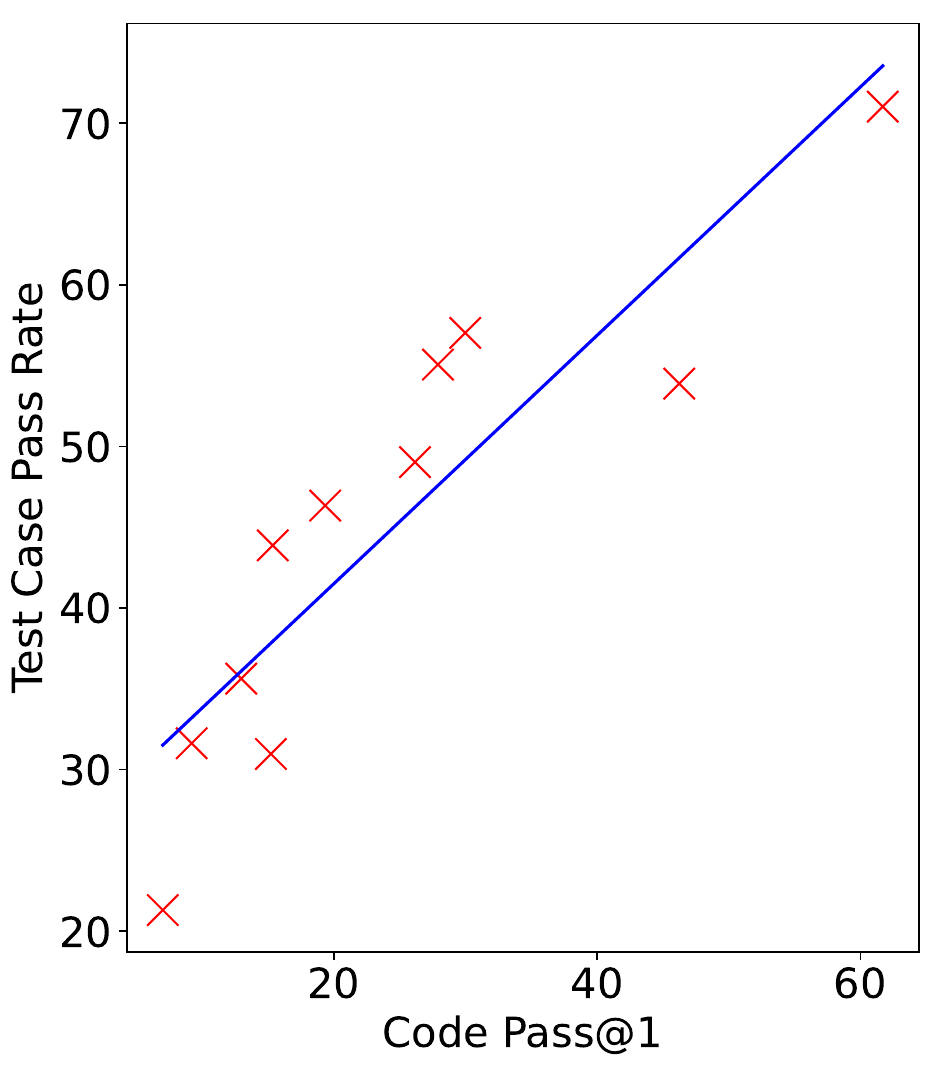}
        \caption{The correlation between code past rate and test pass rate in the ``Oracle'' setting.}
        \label{fig:correlation}
    \end{minipage}\hspace{1em}
    \begin{minipage}{0.6\textwidth}
        \centering
        \includegraphics[width=0.99\linewidth, height=0.23\textheight]{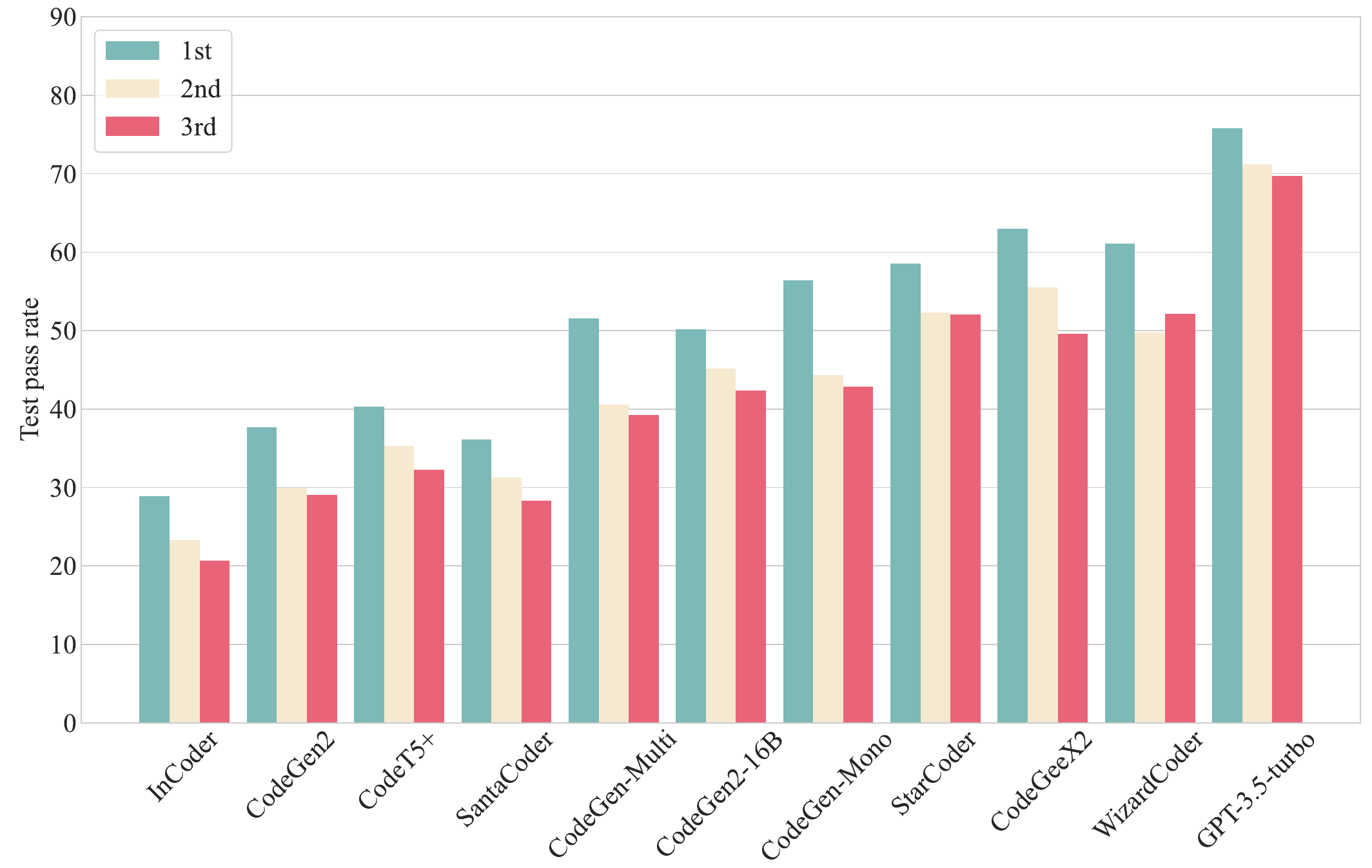} \vskip -0.1in
        \caption{How the correctness of the test cases changes with their order when being generated.} 
        \label{fig:test_case_lines}
    \end{minipage}\vskip -0.13in
\end{figure}

\begin{itemize}

\item \textbf{First}, the test cases generated by LLMs can show a descent pass rate, and this pass rate is even higher than the code pass rate on HumanEval+, which holds for both large and small LLMs. 
Such a result is consistent with intuitions from previous work which rejects code that cannot pass the generated tests to improve the quality of program synthesis.

\item \textbf{Second}, the correctness of the generated test cases is positively correlated with the LLM's ability of generating code (see Figure~\ref{fig:correlation}, where each red cross represents the performance of a model), which means an LLM showing the state-of-the-art program synthesis performance is possibly also the state-of-the-art LLM for program testing.
As shown in Tables~\ref{tab:small model accuracy} and~\ref{tab:large model accuracy}, GPT-3.5-turbo, which synthesizes programs/code with the highest correctness, provides test cases with the highest pass rate (71.03\%) on HumanEval+. 
For an LLM, the more accurate it is capable of synthesizing programs/code on a dataset, the more powerful testing ability will probably be exhibited on the same dataset.
There also exist a few exceptions, e.g., SantaCoder (1.1B) outperforms CodeT5+ (770M) and CodeGen2 (1B) in generating code, but it shows inferior performance in program testing on HumanEval+.
By carefully examining the test cases yielded by SantaCoder on HumanEval+, we found that it tends to generate more complex and longer test cases than CodeT5+ for several problems on HumanEval+, which are often more desirable in program testing. This is also why the SantaCoder test cases show higher coverage rates in Table~\ref{tab:small model accuracy}. 
To be concrete, in Problem 131 in HumanEval+, where the program is required to return the product of all digits with an odd position in a positive integer $n$ (which is the input), the test input provided by CodeT5+ tends to be small for this problem, e.g., $n=2$, while the SantaCoder test cases tend to have more digits (e.g., $n=12358$), which is helpful in digging out hidden bugs. 
Yet, generating longer and more complex test cases is more challenging, and the correctness can be lower.
% Our results in Tables~\ref{tab:small model accuracy} and~\ref{tab:large model accuracy} also show a positive correlation between the obtained branch coverage rate $C$ and the pass rate $P$, and it implies there does not exist much duplication in those correct test cases. 
% For an LLM, if it aims to generate correct and diverse test cases that are also complex, it ought to be very powerful. 

\item \textbf{Third}, as can be seen in Tables~\ref{tab:large model accuracy} and~\ref{tab:model accuracy mbpp}, generating test cases using \emph{large} LLMs with their self-generated code (in the prompts) often leads to a higher level of correctness, compared with the placeholder results. 
This observation is in fact unsurprising, considering that generating code first and test case afterwards resembles the chain-of-thought prompting~\citep{wei2022chain} (if adopting the placeholder is regarded as a plain prompting), which is beneficial to reasoning. 
Moreover, the self-generated performance of an LLM sometimes even outperforms its testing performance with an oracle, and we ascribe this to: 1) randomness in the style of the oracles which are few in number and/or 2) less distribution shift between self-generated code in prompt and the training code, for some powerful LLMs.
% and one may anticipate that generating test cases with more correct function implementations makes more sense.

\item \textbf{Fourth}, with only a few exception, test cases obtained using the oracle code exhibit slightly higher code coverage, while the coverage rate achieved in the other settings (i.e., the self-generated, all-generated, and the placeholder settings) is often slightly lower. 

\end{itemize}

\begin{table}[t]
    \centering
    \small
    \begin{tabular}{c c c c c c c}
    \toprule
    \textbf{Model} & \textbf{Size} & \textbf{Oracle} &  \textbf{Self-generated} & \textbf{All-generated}  & \textbf{Placeholder} \\ %& \textbf{Pass Rate}\\
    \midrule
    InCoder & 1.3B & 21.56\%\,(46.81\%) & 17.98\%\,(46.11\%) & 19.53\%\,(46.45\%) &  22.58\%\,(46.72\%)\\ %& 9.45\\
    CodeGen2 & 1B & 25.61\%\,(54.26\%) & 21.85\%\,(53.09\%) & 23.15\%\,(50.43\%) &  22.81\%\,(52.11\%)\\ %& 10.82\\
    CodeT5+ & 770M & 29.02\%\,(\textbf{56.86\%}) & 24.44\%\,(52.31\%) & 24.84\%\,(\textbf{53.20\%}) & 25.59\%\,(\textbf{55.81\%})\\ %& 14.83 \\
    SantaCoder & 1.1B & \textbf{32.37\%}\,(55.68\%) & \textbf{26.40\%}\,(\textbf{52.38\%}) & \textbf{26.20\%}\,(52.83\%) & \textbf{26.53\%}\,(53.86\%) \\ %& 17.85\\
    \midrule
    % InstructCodeT5+ & 16B & \\
    CodeGen-Multi & 16B & 41.32\%\,(60.63\%) & 35.96\%\,(59.03\%) & 34.17\%,(58.09\%) & 34.84\%\,(58.92\%) \\
    CodeGen2 & 16B & 45.30\%\,(62.15\%) & 38.67\%\,(60.16\%) & 36.77\%\,(58.59\%) & 37.27\%\,(59.16\%) \\
    CodeGen-Mono & 16B & 50.24\%\,(64.39\%) & 43.94\%\,(62.94\%) & 39.55\%\,(61.99\%) & 42.41\%\,(62.31\%) \\
    % StarCoder & 15B & 54.84\% & 46.77\% &  & 48.35\% \\
    StarCoder & 15B & 54.84\%\,(65.10\%) & 46.77\%\,(63.60\%) & 42.80\%\,(61.95\%) & 45.35\%\,(62.66\%) \\
    CodeGeeX2 & 6B & 52.45\%\,(64.64\%) & 44.52\%\,(63.72\%) & 41.72\%\,(60.48\%) & 43.86\%,(63.51\%) \\
    WizardCoder & 15B & 57.85\%\,(\textbf{66.68\%}) & 46.56\%\,(64.86\%) & 41.62\%\,(60.72\%) & 47.45\%\,(64.54\%)\\
    GPT-3.5-turbo & - & \textbf{74.30\%}\,(66.19\%) & \textbf{66.14\%}\,(\textbf{65.30\%}) & \textbf{49.56\%}\,(\textbf{62.95\%}) & \textbf{63.34\%}\,(\textbf{64.72\%})\\
    \bottomrule
    \end{tabular}
    \caption{The pass rates (and coverage rate) of the test cases generated on MBPP.} \vskip -0.1in
    \label{tab:model accuracy mbpp}
\end{table}

The above four takeaway messages can all be inferred from Tables~\ref{tab:small model accuracy},~\ref{tab:large model accuracy}, and~\ref{tab:model accuracy mbpp}. 
% The results mostly hold for the MBPP dataset, as can be seen in Table~\ref{}. 
In addition to all these results, we conduct more experiments to achieve the following takeaway messages.

\begin{itemize}

\item \textbf{Fifth}, by analyzing the relationship between the quality of code in prompts and the correctness of test,
we found that correct code implementation in the prompt often leads to higher quality of test code generation than the case when some incorrect code is given.
We conducted an experiments where we first select programming problems in HumanEval+, where the code pass rate of an LLM is neither 0\% or 100\%. Then we separate self-generated programs/code of the model into two groups, with one group only contains programs/code that are considered as correct and the other only contains incorrect programs/code.
In Table~\ref{tab:correct and incorrect}, we compare the performance of using these two sorts of code in the prompt, for generating test cases using the same LLM.
Apparently, the quality of test cases obtained with correct programs/code is obviously higher.
We further evaluate the overall testing performance of LLMs with only correct self-generated programs/code, if there exists any, in their prompts. 
Unlike in Table~\ref{tab:correct and incorrect} where we do not take problems that can be 100\% or 0\% solved, we take all given problems in this evaluation, except, for every problem, we eliminate all incorrect self-generated programs/code if there exist at least one correct implementation synthesized by the evaluated LLM. 
By doing so, we can observe substantially improved program testing ability on HumanEval+ (i.e., 74.95\% for GPT-3.5-turbo, 56.87\% for WizardCoder, 54.33\% for CodeGeeX2, and 53.24\% for StarCoder), comparing with the original self-generated results in Table~\ref{tab:large model accuracy}. 
The same on MBPP.
Recall that, in our third takeaway message, we have mentioned that test cases obtained with self-generated programs/code sometimes even outperform those yielded in the oracle setting on HumanEval+, maybe partly due to less distribution shift between the synthesized programs/code and the training code, the above results further confirm that, if we can improve the correctness of synthesized programs/code while keeping proper styles, then more powerful testing ability can further be achieved.

% Recall that, in our third takeaway message, we conjecture that testing with self-generated programs/code benefits from less distribution shift, here we further show that, if incorrect code can be eliminated in prompts of the self-generated testing, then it is possible to achieve consistently stronger overall performance even compared with the oracle setting.
% Here we further evaluate overall performance of self-generated testing with only correct (self-generated) program/code implementations in the prompts of LLMs, if there exists any. 
% Unlike in Table~\ref{tab:correct and incorrect} where we do not take problems that can be 100\% or 0\% solved, we take all the problems in this evaluation, except, for every problem, we eliminate all incorrect self-generated programs/code if there exist at least one correct implementation synthesized by the evaluated LLM. 
% By doing so, we observe substantially improved testing ability on HumanEval+ (e.g., 74.95\% for GPT-3.5-turbo, 56.87\% for WizardCoder, 54.33\% for CodeGeeX2, and 53.24\% for StarCoder), 
% % , 30.84\% for SantaCoder, and 33.14\% for CodeT5+)
% comparing with the original self-generated results in Table~\ref{tab:large model accuracy}. 
% The same on MBPP.
% % Tables~\ref{tab:small model accuracy} and~\ref{tab:large model accuracy}. 
% These results confirm that, if we can provide correctly implemented programs/code with proper styles as many as possible, then powerful testing ability can be achieved.

\item \textbf{Sixth}, by conducting an additional experiment, we further compare the quality of test cases collected from different positions in the generation results.
For every set of the three generated test cases, we analyze the relationship between their correctness and the order when they are generated. The results are illustrated in Figure~\ref{fig:test_case_lines}.
As can be seen in the figure, the first generated test case often shows the best correctness and the latterly generated ones are more incorrect. 
This may be due to the fact that the model tends to first generate content with a high level of confidence (which is also more likely to be correct).

\end{itemize}

% \begin{table}[htbp]
%     \centering
%     \small
%     \begin{tabular}{c c c c}
%     \toprule
%     \textbf{Model} & \textbf{Test case 1} & \textbf{Test case 2} &  \textbf{Test case 3}\\ %& \textbf{Pass Rate}\\
%     \midrule
%     InCoder  & 28.87 & 23.29 & 20.69\\ %& 9.45\\
%     CodeGen2  & 37.69 & 29.96 & 29.03\\ %& 10.82\\
%     CodeT5+  & 40.34 & 35.29 & 32.24\\ %& 14.83 \\
%     SantaCoder  & 36.14 & 31.26 & 28.35\\ %& 17.85\\
%     CodeGen-Multi  & 51.55 & 40.60 & 39.25\\ 
%     Codegen2-16B  & 50.14 & 45.13 & 42.40\\ 
%     CodeGen-Mono  & 56.38 & 44.32 & 42.85\\ 
%     StarCoder & 58.53 & 52.30 & 52.06\\ 
%     CodeGeeX2  & 62.99 & 55.54 & 49.56\\ 
%     WizardCoder  & 61.06 & 49.78 & 52.13 \\ 
%     GPT-3.5-turbo  & 75.78 & 71.18 & 69.70 \\ 
%     \bottomrule
%     \end{tabular}
%     \caption{Relationship between generation order and accuracy of test cases}
%     \label{tab:accuracy generation order}
% \end{table}

\section{Improving Program Synthesis Using the Generated Test Cases}

High quality test cases are not only desired in program analyses, but also helpful to program synthesis. 
Previous methods have successfully used generated test cases to improve the performance of LLMs in synthesizing programs/code. 
For instance, \citet{li2023towards} designed a special prompt which involves the test cases as an preliminary, if they are available, for generating programs/code.
\citet{shi2022natural} introduced a Bayes risk decoding mechanism, which executes generated code from a candidate set on a small number of test inputs and selects by marginalizing over implementations that share the same outputs when given these test inputs. It utilizes consistency between the output of code that is correctly implemented.
One step further, \citet{chen2023codet} proposed CodeT, which leverages the LLM to obtain test cases first and tests all synthesized programs/code with these test cases by performing a dual execution agreement, which considers both the agreement between the execution output and the test output and the consistency between the output of correct program implementations, to obtain state-of-the-arts. We encourage interested reader to read the original paper.
% Algo~\citep{zhang2023algo} first prompts LLM to generate an exhaustive program as the reference code and utilize the reference code to generate test cases that can be used to evaluate. 

\begin{table}[!t]
    \centering
    \small
    \begin{tabular}{c c c c c}
    \toprule
    \textbf{Model} & \textbf{Size} & \textbf{w/ correct code} & \textbf{w/ incorrect code}  & \textbf{\#Problem}\\
    \midrule
    % InstructCodeT5+ & 16B & \\
    InCoder & 1.3B & \textbf{28.55\%} & 27.39\% & 27 \\
    CodeGen2 & 1B & \textbf{27.25\%} & 25.74\% & 11 \\
    CodeT5+ & 770M & \textbf{40.19\%} & 36.78\% & 27 \\
    SantaCoder & 1.1B & \textbf{37.45\%} & 34.08\%  & 24 \\
    \midrule
    CodeGen-Multi & 16B & \textbf{55.49\% }& 50.06\% & 32 \\
    CodeGen2 & 16B & \textbf{43.56\%} & 39.31\% & 29 \\
    CodeGen-Mono & 16B & \textbf{45.18\%} & 42.86\% & 56 \\
    StarCoder & 15B & \textbf{58.16\%} & 57.08\%  & 68 \\
    CodeGeeX2 & 6B & \textbf{52.84\%} & 48.63\% & 51 \\
    WizardCoder & 15B & \textbf{48.02\%} & 45.12\%  & 54\\
    GPT-3.5-turbo & - & \textbf{75.39\%} & 68.52\% & 126\\
    \bottomrule
    \end{tabular}
    \caption{With the correct (self-generated) code, the LLMs show stronger ability of generating correct test cases on HumanEval+ (evluated only on those problems that can neither be 0\% solved nor 100\% solved), than in the case where incorrect self-generated code is given in the prompts. Since most LLMs cannot generate any correct code for many hard problems while they often generate incorrect code even for easy problems, the number of tested problems in this experiment increases with the power of the tested LLM, as shown in the rightmost column.} \vskip -0.1in
    \label{tab:correct and incorrect}
\end{table}

In the previous section, we have obtained results about many intriguing properties of the program testing performance of LLMs for code. 
In this section, we would like to drive the readers to think whether it is possible to utilize these results to improve the program synthesis performance, considering that the test cases (hand-crafted and given or automatically generated in particular) are widely and successfully used in program synthesis.
We shall demonstrate that, by utilizing takeaway messages in Section~\ref{sec:main}, the program synthesis performance of previous methods can be improved significantly.
Taking CodeT as an example of the previous state-of-the-art, the method uses a placeholder to generate test cases and treats all the test cases as equally correct as a prior.
However, as discussed in our third takeaway message, using self-generated code helps to achieve more powerful ability in generating correct test cases. 
Moreover, if multiple test cases are provided in a single run of generation given an LLM, the correctness of the test cases decreases with their generation order, as shown in our fifth point.
Hence, to obtain superior program synthesis performance, we introduce two simple modifications to it: 1) we employ the ``self-generated'' setting instead of the ``placeholder'' setting for generating test cases, which means we synthesize programs first and then generate test cases for each program in the prompt, 2) we assign different weights to the generated test cases based on their order in each generation result.

We test the effectiveness of using 1) the prompt which involves self-generated (SG) code and 2) the rank-based weighted (RW) test cases, in improving program synthesis performance on HumanEval+. 
The details of our implementation are introduced as follows. 
Following \citet{chen2023codet}, we used a temperature of 0.8 to generate code and self-generated test cases. 
Each test case is weighted by $p^{i-1}$ with $i$ being its order in the model output, and we let $p=0.8$.

% Then, we use CodeT to test the generated code using these self-generated test cases and placeholder test cases respectively. 
% When determining the weights assigned to test cases, we adopt the first ten training problems from HumanEval+ as our validation set. We set the weight coefficient for the first test case to 1, for the second test case, it's set to some value between 1 and 0, and for the third test example, the weight coefficient is set to some value between the weight coefficient of the second test case and 0. We iterate through weight adjustments with intervals of 0.1 and determine the three weight coefficients that result in the highest code pass rates on the validation set.

Table~\ref{tab:improved code generation} shows the results. 
We compare CodeT with CodeT+SG, CodeT+RW, and CodeT+SG+RW. 
For CodeT, we follow their official implementation and generate $100\times 5$ test cases for each problem.
For fair comparison, we ensure that our solutions with SR and/or RW generate the same numbers of program implementations and test cases as CodeT does. 
Hence, for each problem in HumanEval+, we synthesize a program together with its $5$ test cases for $100$ times when SR and/or RW are incorporated, i.e., we have $i\in\{1,2,3,4,5\}$.
% It can be seen that both SG and WR improves the performance of CodeT considerably, on all LLMs. 
% {\color{red}{It can be seen from the table that both SG and WR improves the program synthesis performance considerably on most LLMs, except for Incoder, CodeGen2-1B, CodeT5+, and SantaCoder for which the test cases generated in the placeholder setting show similar or even higher correctness than in the self-generated setting and SG fails with them.}} 
It can be seen from the table that both SG and WR improves the program synthesis performance considerably on most LLMs, except for Incoder, CodeGen2-1B, CodeT5+, and SantaCoder for which the test cases generated in the placeholder setting show similar or even higher correctness than in the self-generated setting and SG fails with them.
For some LLMs, SG is more powerful, while, on the other models including SantaCoder and StarCoder, RW is more powerful. 
By combining SG and RW, the program synthesis performance of most powerful LLMs in Table~\ref{tab:improved code generation} improves, comparing to only using one of the two.
On GPT-3.5-turbo and WizardCoder, which are the best two models in synthesizing programs on HumanEval+, we achieve +4.22\% and +3.04\% performance gains for CodeT, respectively, with SG \& RW.

We believe there exist other ways of using our takeaway messages for improving the program synthesis performance, and we would like to encourage future work to explore more in this direction.

\begin{table}[!t]
    \centering
    \small
    \begin{tabular}{c c c c c c c}
    \toprule
    \textbf{Model} & \textbf{Size} & \textbf{Baseline} & \textbf{CodeT}  & \textbf{+ SG} & \textbf{+ RW} & \textbf{+ SG \& RW}\\
    \midrule
    % InstructCodeT5+ & 16B & \\
    InCoder & 1.3B & 6.99\% & 9.85\% & 9.45\% & \textbf{10.26\%} & 9.98\% \\
    CodeGen2 & 1B & 9.19\% & 15.15\% & 14.89\% & \textbf{15.67\%} & 15.35\% \\
    CodeT5+ & 770M & 12.95\% & 16.57\% & 16.28\% & \textbf{17.19\%} & 16.98\% \\
    SantaCoder & 1.1B & 15.21\% & 18.43\% & 18.17\% & \textbf{18.75\%} & 18.63\% \\
    \midrule
    CodeGen-Multi & 16B & 15.35\% & 24.50\% & 25.71\% & 25.72\% & \textbf{26.95\%}\\
    CodeGen2 & 16B & 19.33\% & 27.56\% & 28.51\% & 28.43\% & \textbf{29.63\%}\\
    CodeGen-Mono & 16B & 26.15\% & 35.63\% & 36.69\% & 36.63\% & \textbf{37.95\%}\\
    StarCoder & 15B & 27.90\% & 40.46\% & 41.21\% & 42.12\% & \textbf{43.15\%} \\
    CodeGeeX2 & 6B & 29.97\% & 44.16\% & 45.23\% & 44.92\% & \textbf{46.32\%}\\
    WizardCoder & 15B & 46.23\% & 58.41\% & 60.13\% & 59.60\% & \textbf{61.45\%}\\
    GPT-3.5-turbo & - & 61.70\% & 69.25\% & 72.45\% & 70.75\% & \textbf{73.47\%}\\
    \bottomrule
    \end{tabular} \vskip -0.1in
    \caption{\textit{Program synthesis performance} (Pass@1) of LLMs can be significantly improved by using our takeaway messages in Section~\ref{sec:main}. The experiment is on HumanEval+. } \vskip -0.15in
    \label{tab:improved code generation}
\end{table}

\section{Related Work}

\textbf{Test case generation via program analysis. }
Generating reasonable test cases for analyzing programs is a long standing problem in the software engineering community.
Various program analysis techniques, e.g., fuzzing, have been developed for achieving this goal. 
AFL++~\citep{fioraldi2020afl++} is the most popular tool which incorporate many techniques in this category.
A major weakness of these techniques is understandability of the generated test cases.

\textbf{Test case generation via deep learning. }
The invention of transformer~\citep{vaswani2017attention} and self-supervised pre-training~\citep{devlin2018bert, lewis2019bart, raffel2020exploring, radford2018improving} have brought a breakthrough to programming language processing. 
After being trained in a self-supervised manner on a large and diverse code corpus, LLMs have demonstrated remarkable abilities in understanding and synthesizing programs.
We have also witnessed the adaptation of pre-trained LLMs (e.g., ChatGPT) to fuzz testing~\citep{xia2023universal} very recently. 
Nevertheless, there still lack and requrie in-depth analyses and intensive comparisons of different LLMs in program testing. 
In particular, powerful LLMs emerge continuously.
For instance, the recent WizardCoder~\citep{luo2023wizardcoder} exhibits an obivous program synthesis superiority over other open-source LLMs. 
In our study, we focus on the analyses and comparison of the LLMs in writing test code and generating test cases.
% % Extensive research on test case generation has been carried out on transformer-based language models. 
% % AthenaTest~\citep{tufano2020unit} generate human-written tests from source code by treating test case generation as a translation task and finetunes the model on the Method2Test~\citep{tufano2022methods2test} dataset. A3Test~\cite{alagarsamy2023a3test} improves pretraining by using a PLBART model. and verifies naming and test signatures consistency. 
% CodeT~\citep{chen2023codet} directly sample test cases from powerful code generation models like Codex in the zero-shot setting with simple prompts. And like them, we use code language model to generate test cases.

\textbf{Evaluation of Large Language Model. }
Recently, large language models (LLMs) has incited substantial interest in both academia and industry. In order to evaluate the capabilities of large language models, a variety of effort have been devoted from the perspectives of natural/programming language processing accuracy, robustness, ethics, biases, and trustworthiness, etc. For instance, PromptBench \citep{zhu2023promptbench} demonstrates that current LLMs are sensitive to adversarial prompts, and careful prompt engineering is necessary for achieving descent performance with them. 
Another example, DecodingTrust~\citep{wang2023decodingtrust}, offers a multifaceted exploration of trustworthiness of the GPT models, especially GPT-3.5 and GPT-4. 
The evaluation expands beyond the typical trustworthiness concerns to include several new critical aspects. 
Agentbench~\citep{liu2023agentbench} evaluates LLM as agents on challenging tasks in interactive environments. 
Their experimental results show that, while top commercial LLMs present a strong ability of acting as agents in complex environments, there is a significant disparity in performance between them and their open-source competitors.

\section{Conclusion}

In this paper, we have performed thorough analyses of recent LLMs (mostly LLMs for code) in testing programs/code. 
Through comprehensive experiments with 11 LLMs on programming benchmark datasets including HumanEval+ and MBPP (the sanitized version), we have uncovered a range of intriguing characteristics of these LLMs for program/code testing.
We have illustrated how the program testing capabilities of these LLMs can be enhanced in comparing intensive empirical results in four different settings. 
Based on our findings, we are also capable of improving the performance of state-of-the-art LLMs in synthesizing programs/code with test cases of higher quality. 
As a preliminary research work, we believe our paper can provide new research insights and spark new ideas in program/code synthesis, test-case generation, and LLM understanding, and we look forward to future exploration in this direction in future work.

\bibliography{iclr2024_conference}
\bibliographystyle{iclr2024_conference}

\appendix
\section{Appendix}
% \subsection{Why does SantaCoder perform worse than CodeT5+ in generating test cases on the HumanEval+ dataset?}
\subsection{Further Analysis of Experimental Results}
In this part, we provide further analysis of the experimental results in Section \ref{sec:main}. 

With regard to the situation where the test case quality generated by SantaCoder is lower than that generated by CodeT5+ on the HumanEval+ dataset, we have explained that this is probably because SantaCoder tends to generate longer and more complex test cases. 
Here we further demonstrate that SantaCoder is capable to generate more accuracy output when given the same testing input as that of CodeT5+'s.
To show this, we first extract the input part of the test cases (which includes testing inputs paired with their corresponding outputs) generated by CodeT5+ in the oracle setting. 
We then let SantaCoder to generate testing outputs given these inputs, and assessed the accuracy of such test cases.
% To demonstrate that SantaCoder's ability to generate test cases is superior to CodeT5+, i.e., the correctness of the generated test cases is positively correlated with the LLM's ability to generate code, we extract the input of the first test case generated by CodeT5+ under the oracle setting and have SantaCoder generate the corresponding output. We adopt the same experimental setting as in Section \ref{sec:main}. 
The results show that, given these testing inputs already, SantaCoder and CodeT5+ obtain an correctness of \textbf{41.67\%} and \textbf{40.34\%}, respectively, showing that SantaCoder is indeed stronger, if the same testing input is given and it does not have the chance to yeild more complex testing inputs.

\end{document}